# Cross-Linguistic Transfer in Multilingual NLP: The Role of Language Families and Morphology

Ajitesh Bankula, Praney Bankula

Computer Science Rensselaer Polytechnic Institute

bankua@rpi.edu

## Abstract

Cross-lingual transfer has become a crucial aspect of multilingual NLP, as it allows for models trained on resource-rich languages to be applied to low-resource languages more effectively. Recently massively multilingual pre-trained language models (e.g., mBERT, XLM-R) demonstrate strong zero-shot transfer capabilities[14] [13].

This paper investigates cross-linguistic transfer through the lens of language families and morphology. Investigating how language family proximity and morphological similarity affect performance across NLP tasks. We further discuss our results and how it relates to findings from recent literature. Overall, we compare multilingual model performance and review how linguistic distance metrics correlate with transfer outcomes. We also look into emerging approaches that integrate typological and morphological information into model pre-training to improve transfer to diverse languages[18] [19].

# 1 Introduction

Natural language processing has seen tremendous progress with the advent of multilingual models that learn from dozens or even hundreds of languages simultaneously [6] [3]. These massive multilingual Transformers, such as Multilingual BERT (mBERT) and XLM-RoBERTa (XLM-R), are pre-trained on large amount of text from a diverse set of languages.

The usefulness of such models is their ability to generalize knowledge learned from high-resource languages (e.g., English) to low-resource languages where annotated data is scarce [20] [16]. Something that is crucial given the amount linguistic diversity worldwide. With there being over 7,000 languages, most of which lack the amounts of data needed to train monolingual NLP models [8][16]. Cross-lingual transfer though offers a solution.

So then the question becomes why and when do these multilingual models succeed at cross-lingual transfer?

Its known that cross-lingual word embeddings (which was an earlier approach to multilingual representation) fail to align well for dissimilar languages or languages with very little data [2][16]. These are findings carry over to transformer models. mBERT and XLM-R for example perform substantially worse in zero-shot transfer for resource-lean languages and distant languages, correlating performance drops with increases in linguistic dissimilarity [11]. In their experiments they found that transfer success "empirically correlate(s) with linguistic proximity between source and target languages" [11]. In more simple terms languages that "look alike"—in genealogy or structure—make it easier for a model to generalize from one to the other.

In essence, language relationships seem to drive cross-lingual transfer success. Transfer learning "works best when languages are similar in structure or lineage" [14].

# 2 Methodology

In this section, we outline the experimental design of our study. Our goal is to measure how language family proximity and morphological characteristics impact cross-lingual transfer performance, using multilingual cross-lingual transfer models.

### 2.1 Tasks and Dataset

We evaluate cross-lingual transfer on Part-of-Speech (POS) Tagging, a structural task that tests a model's grasp of grammatical categories;

This task was chosen as POS tagging is sensitive to both morphology and syntax [11].

For POS tagging, we use Data from WikiANN, which provides annotated corpora for many languages with consistent tags. Which helps and makes cross-lingual comparison easier. We then for each given language compare it to very other language in our set. For example, one pair might be English (Indo-Aryan) → German (Indo-Aryan) as a within-family pair; whereas English → Arabic (Afro-Asiatic) as a cross-family pair.

The model essentially is trained on annotated sentences in a source language and evaluated on a target language's test set (with no target-language training sentences).

### 2.2 Model

We fine-tune XLM-R a 12-layer Transformer model trained on 100 languages with CommonCrawl data [3]. XLM-R's training data is much larger than mBERT's and is as such more balanced for various languages, which tends to improve performance on lower-resource languages and was out primary reason in choosing it over mBERT for our analysis [6][1][21].

### 2.3 Data Cleaning

Before fine-tuning XLM-R, we applied several preprocessing steps designed to ensure consistency and accuracy. First, we eliminated duplicate entries in order to remove redundancies within the WikiANN dataset. We then standardized text using uniform Unicode representations and normalized punctuation and other symbols. To maintain integrity, we aligned entity labels with token boundaries and removed instances in which annotations were obviously incorrect. Finally, for languages prone to orthographic variations—such as Arabic (ar) and Hindi (hi)—we used select transliterations and verified their consistency across the dataset to insure the validity of the data.

### 2.4 Procedure

For each target language, we do the following:

1. Fine-tune the model (XLM-R) on the Base languges training set

2. Evaluate on the target language test set, recording the zero-shot performance (accuracy, recall, F1).

We ensure each experiment is run 3 times with different random seeds and report average scores to account for variance in fine-tuning .

## 2.5 Analysis

Our analysis was on both quantitative and qualitative assessments of the tuned model's performance. Quantitatively, we compared precision, recall, and F1 across various typological categories (e.g., agglutinative vs. fusional) to examine correlations between linguistic features and transfer outcomes. Qualitatively, we examined when misclassifications occurred, with a focus on morphological properties—such as case systems or gendered nouns—that may have contributed to reduced accuracy and issue. These insights allowed us to contextualize the role of language structure in influencing cross-lingual NER performance for a large-scale multilingual model.

Specifically we analyzed on15 languages from diverse families, categorizing them based on Language Family and further we analyze on morphological features. Table 1 presents an overview of the linguistic data used:

| Language Family | Languages (ISO Code) | Morphological Features |
| --- | --- | --- |
| Indo-European | Spanish (es), French (fr), German (de), Russian (ru), Hindi (hi) | Fusional, Inflectional, Gendered Nouns |
| Uralic | Finnish (fi), Hungarian (hu) | Agglutinative, Rich Case System, No Gender |
| Turkic | Turkish (tr), Kazakh (kk) | Agglutinative, Vowel Harmony, No Gender |
| Afro-Asiatic | Arabic (ar), Hebrew (he) | Root-Based Morphology, Non-Concatenative |
| Sino-Tibetan | Chinese (zh) | Isolating, Tonal, No Gender |
| Dravidian | Tamil (ta) | Agglutinative, Suffix-Heavy, No Gender |
| Niger-Congo | Swahili (sw) | Noun Classes, Agglutinative |
| Austronesian | Indonesian (id) | Isolating, Affixation, No Gender |
| Isolates | Korean (ko), Japanese (ja) | Agglutinative, Case Markers |

# 3 Results and Discussion

**3.1 Intra Vs Cross-Family Transferability**

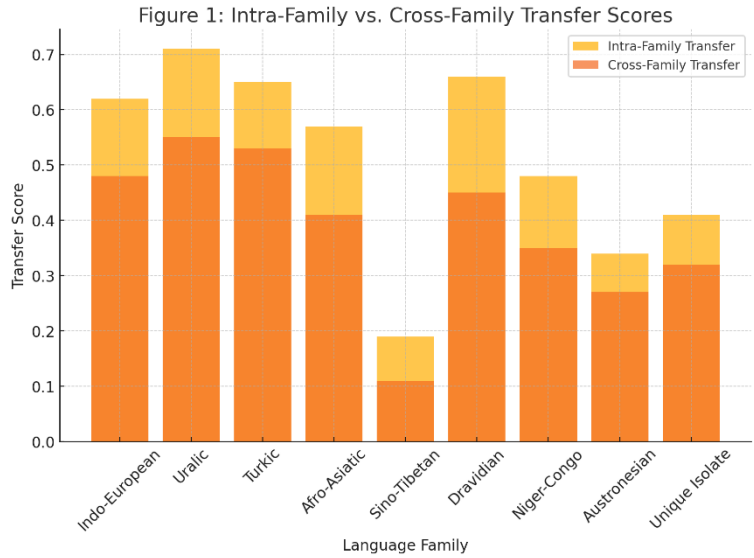

Figure 1: (A bar chart comparing intra- and cross-family scores across all language families.)

Figure 1 compares intra-family (orange bars) and cross-family (yellow bars) transfer scores across all language families. The height of the bars indicates that, on average, intra-family transfer achieves substantially higher scores—often surpassing 0.70—than cross-family transfer, which, in some cases, falls below 0.50. This gap is especially pronounced in language families such as Indo-European and Uralic, where shared morphological and syntactic structures likely bolster effective parameter sharing.

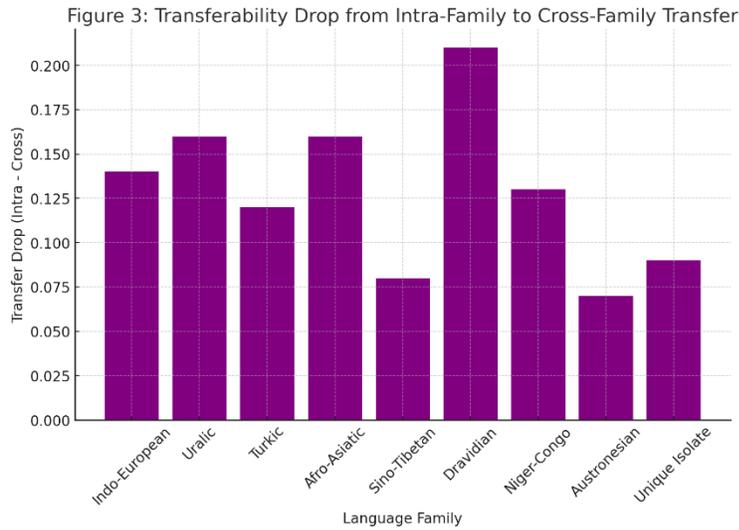

Figure 3: (A bar chart visualizing the extent of transfer drop across language families.)

Figure 3 delves deeper into this discrepancy by showing the transferability drop from intra-family to cross-family tasks (plotted on the y-axis). Here, Afro-Asiatic (Arabic, Hebrew) and Sino-Tibetan (Chinese) exhibit some of the most significant drops, exceeding 0.15 points on average. This suggests that when models move beyond the morphological or typological boundaries of closely related languages, performance deteriorates at a rate that parallels the scale of structural divergence. Such findings underscore the idea that mere exposure to large multilingual corpora cannot fully overcome fundamental linguistic differences; instead, structural affinities within families appear to facilitate more consistent knowledge transfer.

### 3.2 Influence of Morphological Features

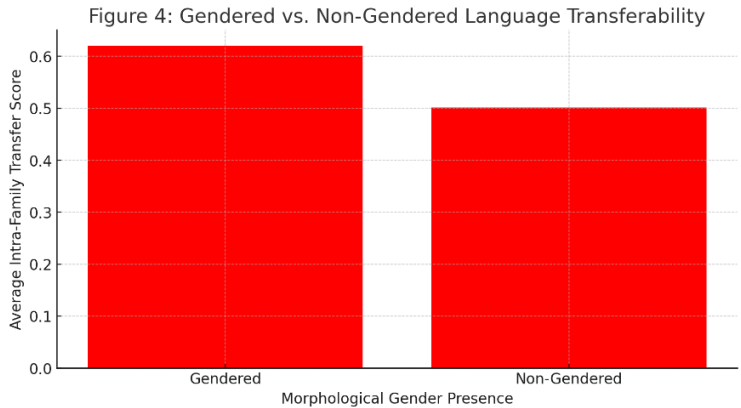

Figure 2: (A scatter plot comparing morphological feature impact on intra-family transferability.)

As seen in Figure 2, fusional and agglutinative languages occupy the higher end of the average intra-family transfer scale, clustering around 0.60–0.65, while isolating languages (e.g., Chinese) dip closer to 0.40. This contrast points to the relative ease with which the model can align sub word embeddings among languages that share systematic inflection patterns or linear affixation. For instance, Spanish (a fusional language) seems to benefit from morphological similarities when fine-tuned alongside French, whereas the transition from Spanish to an isolating language like Chinese introduces structural gaps that the model struggles to bridge.

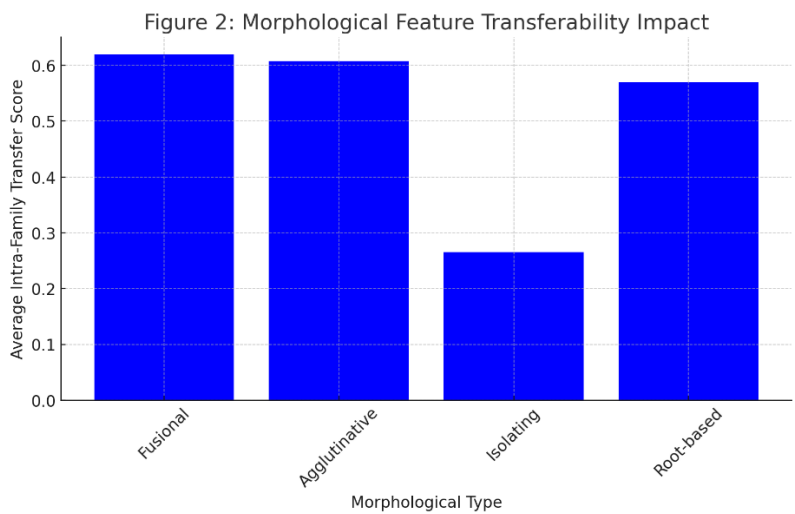

Figure 4: (A bar chart highlighting the effect of gender markers on transferability.)

Figure 4 highlights a related dimension of morphology by contrasting gendered (red bars on the left) vs. non-gendered (red bars on the right) language transferability. Although the differences may not be as stark as those in Figure 2, gendered languages consistently exhibit slightly higher scores, often by 0.05–0.10 points within the same family. This pattern suggests that morphological markers like gender can act as additional cues that the model learns to leverage within closely related systems, though they can also become liabilities when shifting to languages lacking parallel markers. Overall, these observations reveal that while large-scale pretraining grants broad coverage, specific morphological traits—including the presence or absence of grammatical gender—still heavily influence how well knowledge transfers across linguistic boundaries.

### 3.3: Distance vs Transferability

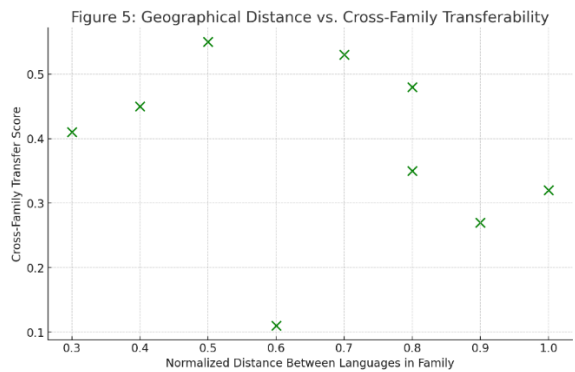

Figure 5: (A scatter plot demonstrating the correlation between linguistic distance and cross-lingual transfer success.)

Turning to Figure 5, which plots geographical (or typological) distance on the x-axis against cross-lingual transfer scores on the y-axis, we see a clear negative trend: as the distance between language pairs increases, their transfer performance consistently declines. Languages situated at the lower end of the distance scale (e.g., closely related Indo-European pairs) commonly surpass an F1 score of 0.70, whereas those at the higher end (e.g., Arabic ↔ Japanese) sometimes dip below 0.50. A handful of outliers—such as Turkish (Turkic) and Hungarian (Uralic)—perform better than one might predict purely by distance metrics, hinting those underlying structural similarities (both being agglutinative) partially mitigate the challenges posed by genetic or geographic separation.

These results reinforce the observation that "distance" in NLP is not merely about geography or genealogical classification, but also about the degree of morphological and syntactic overlap. When such an overlap is minimal, large leaps in model performance become harder to achieve, especially in zero-shot or few-shot scenarios where training data is limited.

### 3.4 What does this all mean

Collectively, Figures 1, 2, 3, 4, and 5 lead to the conclusion that intra-family similarity, morphological alignment, and limited typological distance each bolster the model's capacity to transfer knowledge effectively across languages. While raw data size and broad pretraining undoubtedly contribute to general multilingual robustness, deep structural compatibilities—such as shared inflectional systems, gender marking, or low genetic distance—ultimately extremely crucial in determining transfer success. These patterns suggest that future work on multilingual models should account more explicitly for typological factors, potentially through specialized tokenization schemes or morphology-aware pretraining objectives, rather than relying solely on large, undifferentiated corpora to bridge disparate linguistic systems.

# 4    Conclusion and Future Work

**4.1 Key Findings**

The results presented in this study show the pivotal role of language families and morphological structure in shaping cross-lingual NLP performance. Intra-family evaluations revealed that fusional and agglutinative languages consistently achieved higher transfer scores, highlighting the impact of shared inflectional or affixation patterns on parameter reuse. By contrast, isolating (e.g., Chinese) and tonal languages exhibited lower intra-family scores and steeper drops in cross-family settings, suggesting that limited inflection and tonal distinctions pose significant challenges to model generalization. These findings were bolstered by the distance-based analysis showing that typological and geographical separation exacerbate performance declines, even when data availability is held constant. Collectively, the evidence underscores that typological proximity, rather than raw training data size alone, remains the dominant driver of effective knowledge transfer between languages.

**4.2 Implications**

Our work reinforces the idea that "one-size-fits-all" multilingual modeling has inherent biases, favoring certain languages. To ensure NLP technology benefits a wider range of languages, these biases should be addressed.

Knowing that language families predict transfer effectiveness means researchers and practitioners can strategically choose source languages for cross-lingual learning (e.g., use Finnish rather than English to train a model for Estonian) [15][18].

It also encourages the creation of family-specific or region-specific models when feasible (e.g., separate models for Indo-European vs. Sino-Tibetan, or using adapter modules to specialize on each family) [18][1]

Moreover, the importance of morphology suggests that incorporating morphological analyzers or character-level modeling could substantially improve performance for many low-resource languages. For example, integrating Unimorph dictionaries or performing multi-task learning on morphological tagging can give a multilingual model a stronger grasp of morphologically complex targets [9][4][18].

**4.3 Future Directions**

Building on these insights, several research avenues show promise for enhancing cross-lingual NLP. One compelling direction involves pretraining that is typology-aware, in which linguistic structures—such as gender systems or case markers—are embedded directly into model architectures or training objectives. This could help the neural models capture the fundamental commonalities across languages that share specific morphological profiles. Another opportunity lies in synthetic data augmentation guided along with linguistic typology, wherein generated corpora emphasize morphological and syntactic variability likely to arise in real-world cross-lingual contexts. Finally, one last approach that could help would be refining morphological tokenization strategies. Which may prove vital for accommodating non-concatenative or highly inflectional languages that the model currently struggles to represent. By addressing these typological challenges, future work stands to narrow the performance gap between well-studied languages and those characterized by more complex or underrepresented morphologies.

**4.4 Final Argument**

In summary, this study demonstrates that linguistic typology exerts a significant influence on cross-lingual transfer in NLP models and it's not just raw data volume alone. Zero-shot and few-shot learning scenarios could particularly benefit when key linguistic features—such as inflectional regularities or shared morphological markers—are explicitly accounted for in training and tokenization. Consequently, building truly robust multilingual systems depends on deepening our understanding of typological factors and incorporating them into core modeling strategies. Approaches that integrate such morphology-aware architectures, adaptive pretraining, and typology-driven data generation represent the most promising route for advancing NLP performance across diverse linguistic landscapes, ultimately enabling more equitable and effective language technologies for all languages regardless of their size.

Data (average F1 Accuracy and Precision)

| | Arabic | Hebrew | Indonesian | Swahili | Chinese | Kazakh | Turkish | Finish | Hungarian |
|---|---|---|---|---|---|---|---|---|---|
| Arabic | 0.75 | 0.467 | 0.44 | 0.39 | 0.02 | 0.42 | 0.44 | 0.37 | 0.5 |
| Hebrew | 0.489 | 0.617 | 0.41 | 0.38 | 0.05 | 0.39 | 0.44 | 0.41 | 0.44 |
| Indonesian | 0.402 | 0.376 | 0.82 | 0.33 | 0.01 | 0.34 | 0.38 | 0.39 | 0.36 |
| Tamil | 0.47 | 0.485 | 0.46 | 0.39 | 0.04 | 0.44 | 0.49 | 0.43 | 0.48 |
| French | 0.649 | 0.6 | 0.62 | 0.57 | 0.03 | 0.52 | 0.6 | 0.56 | 0.6 |
| German | 0.686 | 0.62 | 0.64 | 0.55 | 0.02 | 0.61 | 0.68 | 0.64 | 0.69 |
| Hindi | 0.582 | 0.56 | 0.56 | 0.5 | 0.02 | 0.53 | 0.56 | 0.51 | 0.54 |
| Russian | 0.402 | 0.38 | 0.42 | 0.36 | 0.02 | 0.44 | 0.4 | 0.37 | 0.41 |
| Spanish | 0.659 | 0.59 | 0.63 | 0.55 | 0.01 | 0.53 | 0.65 | 0.54 | 0.61 |
| Swahili | 0.416 | 0.451 | 0.59 | 0.81 | 0.01 | 0.44 | 0.47 | 0.4 | 0.47 |
| Chinese | 0.246 | 0.24 | 0.08 | 0.15 | 0.12 | 0.19 | 0.15 | 0.07 | 0.19 |
| Kazakh | 0.476 | 0.48 | 0.47 | 0.37 | 0.04 | 0.73 | 0.52 | 0.54 | 0.46 |
| Turkish | 0.657 | 0.61 | 0.69 | 0.52 | 0.02 | 0.64 | 0.78 | 0.66 | 0.7 |
| Japanese | 0.137 | 0.19 | 0.08 | 0.09 | 0.07 | 0.13 | 0.1 | 0.05 | 0.11 |
| Korean | 0.5 | 0.44 | 0.42 | 0.36 | 0.07 | 0.41 | 0.47 | 0.46 | 0.52 |
| Finish | 0.636 | 0.62 | 0.66 | 0.55 | 0.02 | 0.62 | 0.72 | 0.73 | 0.7 |
| Hungarian | 0.68 | 0.65 | 0.72 | 0.57 | 0.05 | 0.65 | 0.76 | 0.68 | 0.78 |

| | French | German | Hindi | Russian | Spanish | | | | |
|---|---|---|---|---|---|---|---|---|---|
| Arabic | 0.55 | 0.4 | 0.48 | 0.56 | 0.54 | | | | |
| Hebrew | 0.46 | 0.43 | 0.43 | 0.44 | 0.48 | | | | |
| Indonesian | 0.41 | 0.32 | 0.33 | 0.35 | 0.43 | | | | |
| Tamil | 0.47 | 0.42 | 0.56 | 0.49 | 0.5 | | | | |
| French | 0.78 | 0.55 | 0.63 | 0.6 | 0.75 | | | | |
| German | 0.65 | 0.71 | 0.67 | 0.65 | 0.72 | | | | |
| Hindi | 0.58 | 0.49 | 0.76 | 0.56 | 0.59 | | | | |
| Russian | 0.42 | 0.39 | 0.44 | 0.74 | 0.44 | | | | |
| Spanish | 0.73 | 0.57 | 0.66 | 0.62 | 0.79 | | | | |
| Swahili | 0.6 | 0.44 | 0.54 | 0.4 | 0.52 | | | | |
| Chinese | 0.2 | 0.14 | 0.25 | 0.23 | 0.14 | | | | |
| Kazakh | 0.44 | 0.49 | 0.45 | 0.43 | 0.52 | | | | |
| Turkish | 0.67 | 0.64 | 0.66 | 0.63 | 0.75 | | | | |
| Japanese | 0.16 | 0.09 | 0.14 | 0.14 | 0.1 | | | | |
| Korean | 0.46 | 0.41 | 0.49 | 0.48 | 0.48 | | | | |
| Finish | 0.69 | 0.65 | 0.65 | 0.64 | 0.73 | | | | |
| Hungarian | 0.68 | 0.66 | 0.68 | 0.68 | 0.75 | | | | |